\PassOptionsToPackage{round,authoryear}{natbib}
\PassOptionsToPackage{table,dvipsnames}{xcolor}
\documentclass[]{style/company_light} 

\usepackage[utf8]{inputenc}
\usepackage{hyperref}
\usepackage{url}
\usepackage{booktabs}
\usepackage{amsfonts}
\usepackage{amsmath}
\usepackage{nicefrac}
\usepackage{microtype}
\usepackage{graphicx}
\usepackage{subcaption}
\usepackage{multirow}
\usepackage{enumerate}
\usepackage{enumitem}
\usepackage[most]{tcolorbox}
\usepackage{adjustbox}
\usepackage{array}
\usepackage{amssymb}
\usepackage{float}
\microtypesetup{expansion=false}
\NewDocumentCommand{\todo}
{ mO{} }{\textcolor{magenta}{\textsuperscript{\textit{TODO}}\textsf{\textbf{\small[#1]}}}}

\newcommand{\ph}[1]{\textcolor{magenta}{\underline{\textsf{#1}}}}

\newcommand{\figorph}[3]{\IfFileExists{#1}{\includegraphics[width=#2]{#1}}{\fbox{\parbox[c][0.28\linewidth][c]{0.95\linewidth}{\centering\ph{Figure pending: #3}}}}}

\newcommand{\BenchmarkName}{Quant-Arena}

\newcommand{\ModelClaudeOpus}{Claude~Opus~5.5}
\newcommand{\ModelDeepSeek}{DeepSeek~V4.1~Flash}
\newcommand{\ModelGemini}{Gemini~3.8~Flash}
\newcommand{\ModelClaudeSonnet}{Claude~Sonnet~5.5}
\newcommand{\ModelSol}{GPT-6~Sol}
\newcommand{\ModelTerra}{GPT-5.6~Terra}
\newcommand{\ModelAstra}{GPT-6~Astra}
\newcommand{\ModelGLM}{GLM-5.3}
\newcommand{\ModelKimi}{Kimi~K3}

\newcommand{\modellogo}[1]{\raisebox{-0.12em}{\includegraphics[width=1em,height=1em,keepaspectratio]{assets/model-logos/#1.pdf}}\hspace{0.35em}}

\title{Quant-Arena: Benchmarking LLM Agents for Quantitative Trading Research}

\author{Houde Qian}
\author{Zheng Jiang}
\author{Wenyi Li}
\author{Chenxi Xu}
\author{Xinjie Lin}
\author{Yuzhao Peng}
\author{Ruoyu Han}
\author{Youjie Zheng}
\author{Kaisen Yang}
\author[*]{Qinhuai Na}
\author[*]{You Fu}
\author[*]{Calvin Xiao}
\affiliation{Navers Lab, Einsia.AI}
\contribution[*]{Corresponding authors.}

\abstract{%
Quantitative research is an iterative loop: a researcher proposes a market hypothesis, implements it as an executable signal, tests it on historical data, and decides what to keep. Existing benchmarks score formulas, code, or individual trading decisions, leaving open whether large language models (LLMs) can complete this loop and produce strategies that work beyond the research period. We introduce \textbf{\BenchmarkName}, a benchmark and public leaderboard for end-to-end quantitative research on real A-share market data. A model constructs features and stock-scoring code, receives feedback only on 2022--2023, and freezes one strategy for assessment on the sealed year 2024. A cash-account evaluator applies trading constraints and costs, including market impact estimated from a simulated order book. Scores are dominated by realized excess return and complemented by signal quality and the diversity of features used by the strategy. We evaluate nine frontier models, among which \ModelAstra\ scores highest; yet no model beats the equal-weight market in 2024. \BenchmarkName\ is open to submissions from any model or agent harness.
}

\date{\today}
\metadata[HomePage]{\url{https://lab.einsia.ai/quant-arena}}

\begin{document}

\maketitle

\vspace*{-0.6em}\noindent\begin{minipage}{\linewidth}
\centering
\includegraphics[width=0.9\linewidth]{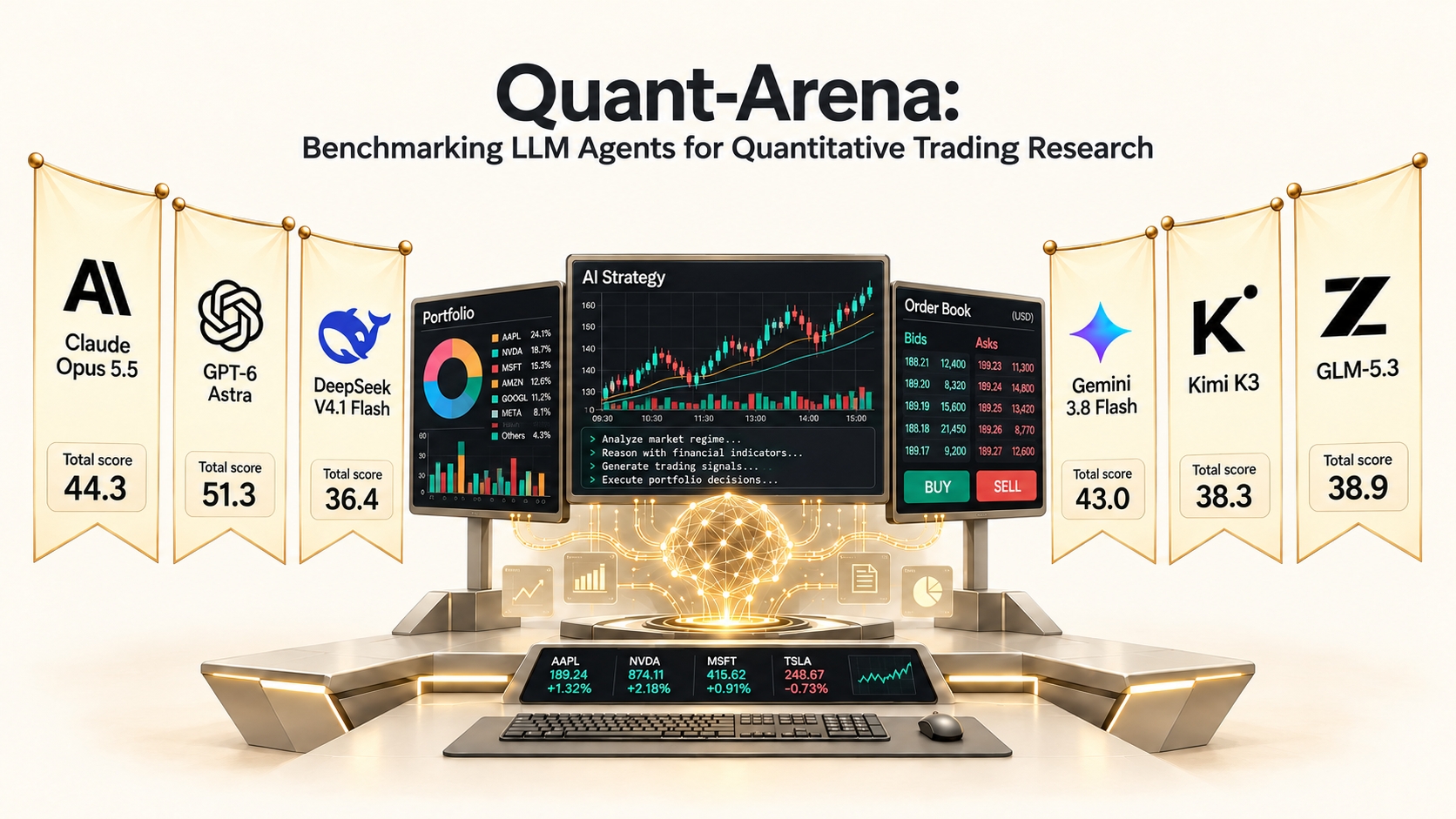}
\captionof{figure}{\BenchmarkName\ evaluates LLMs on complete quantitative research runs. Banners show total scores on the iterative research track.}
\label{fig:teaser}
\end{minipage}
\clearpage

\section{Introduction}
\label{sec:intro}

Quantitative research turns hypotheses about market behavior into executable trading strategies. Researchers construct features from market observations, combine them into predictive factors or signals, and use those signals to build and test portfolios. Large language models (LLMs) are a natural fit for this research loop: they can propose hypotheses, implement them in code, interpret feedback, and revise their approach. Recent systems already explore this potential through interactive alpha discovery and iterative factor search \citep{Wang2023AlphaGPT,Tang2025AlphaAgent,Han2026QuantaAlpha}.

Existing benchmarks measure different parts of this workflow. AlphaQT-Bench evaluates the correctness of quantitative trading code, while AlphaBench studies the generation, evaluation, and search of formulaic factors \citep{AlphaQTBench2026,Luo2026AlphaBench}. StockBench evaluates agents that make sequential trading decisions from market information \citep{StockBench2025}. AlphaForgeBench extends evaluation to generating factor and strategy code from natural-language objectives and backtesting it on historical daily market data \citep{Zhang2026AlphaForgeBench}.
To our knowledge, this is the first benchmark on real market data in which models construct features from raw order and trade records, revise their signals using bounded feedback, and submit a frozen strategy for out-of-sample assessment under common trading rules. This setting evaluates feature construction, signal combination, and strategy selection through their joint effect on the final portfolio.

\begin{figure*}[t]
    \centering
    \figorph{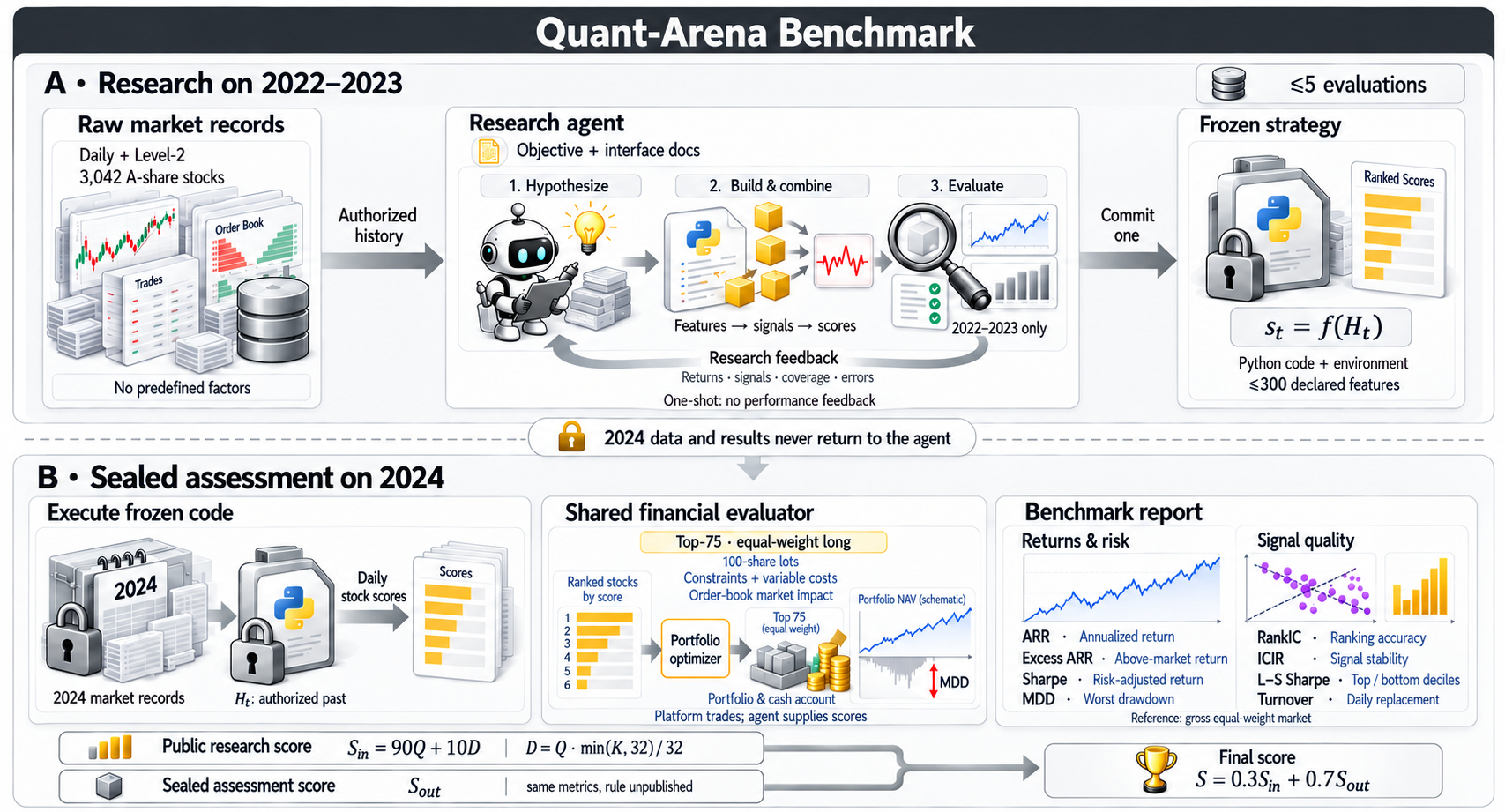}{\linewidth}{Benchmark pipeline}
    \caption{A model constructs features and stock-scoring code from raw market observations, revises candidates using research-period feedback, and freezes one strategy for assessment on the sealed period under common portfolio and execution rules.}
    \label{fig:pipeline}
\end{figure*}

We introduce \textbf{\BenchmarkName}, a benchmark for end-to-end quantitative research on real A-share market data (Figure~\ref{fig:pipeline}). At the input end, model-generated Python programs access raw Level-2 records, including individual orders and trades and order-book snapshots, through a controlled data interface. Models determine which features to construct and how to combine them into daily stock scores. During research, they can evaluate candidate strategies on 2022--2023, inspect the permitted feedback, and revise their code within a fixed budget. Each run then commits one frozen strategy. At the output end, the platform executes that strategy on the sealed year 2024 and measures the portfolio's returns after trading costs.

\BenchmarkName{} applies common portfolio and execution rules to make these outcomes comparable. Daily scores determine an equal-weight portfolio of the highest-ranked stocks, and a cash-account evaluator applies lot-size constraints and order-dependent costs, including market impact estimated from a simulated order book. Out-of-sample return and excess return over the market provide direct measures of profitability. Risk and signal statistics and the diversity of features used by the strategy provide complementary views of the result. Together, these measurements distinguish finding a predictive signal from turning it into a profitable portfolio.

A common research budget controls how much search each run can undertake. Repeated backtests favor candidates that fit historical noise \citep{Bailey2017Backtest}, so the selected strategy must be assessed on data that its research never observed. The platform also records candidate evaluations and final submissions. These records allow us to examine how models use feedback and select strategies alongside their eventual financial outcomes.

Our contributions are:

\begin{itemize}[leftmargin=*,itemsep=2pt]
    \item \textbf{An end-to-end quantitative research benchmark.} \BenchmarkName{} connects feature and signal construction from real market events to the out-of-sample performance of an executable strategy, making the complete research run the unit of evaluation.
    \item \textbf{A common research and financial evaluation protocol.} A bounded research process, frozen strategy submission, and sealed assessment under shared portfolio and trading rules support comparisons of profitability, risk, signal quality, and feature diversity.
    \item \textbf{An empirical analysis of LLM quantitative research.} We evaluate nine frontier models. No model beats the equal-weight market in 2024: the stocks a model ranks highest typically underperform the market average. The strongest model gains mainly by learning when to stay out of the market rather than by selecting better stocks.
    \item \textbf{A public evaluation platform.} \BenchmarkName{} is released as a public arena that accepts submissions from any model or agent harness under the same protocol and sealed assessment.
\end{itemize}

The public leaderboard also lists Kepler-1, a proprietary strategy developed by Einsia AI.

\section{The \BenchmarkName\ Benchmark}
\label{sec:benchmark}

\BenchmarkName{} evaluates the out-of-sample profitability of stock-selection strategies developed by models from historical market data.

\subsection{Task Formulation}
\label{sec:task}

A model receives a research objective, market-data access, and documentation for the data and evaluation interfaces. Within a fixed research budget, it writes a strategy $f$ that constructs features, combines them into predictive signals, and scores stocks. For each trading day $t$, the strategy computes

\[
\mathbf{s}_t=f(\mathcal{H}_t),
\]

where $\mathcal{H}_t$ contains the authorized historical observations available at that time and $\mathbf{s}_t$ contains stock scores. The model determines the features and scoring function. A common financial evaluator converts these scores into portfolio positions and returns. The final submission is the frozen strategy program $f$.

\subsection{Market Data and Benchmark Construction}
\label{sec:data}

We use real market observations from a fixed data release, stock universe, and trading calendar. Models construct features from the raw inputs without a predefined factor library. Missing observations and delisted stocks are retained without redefining daily universe membership.

\begin{table}[htbp]
\centering
\small
\caption{Market data and temporal split used by the benchmark.}
\label{tab:benchmark_data}
\begin{tabular}{@{}p{0.24\linewidth}p{0.70\linewidth}@{}}
\toprule
Property & Specification \\
\midrule
Stock universe & Selected universe of 3,042 stocks \\
Inputs & Daily market data and raw Level-2 quotes, orders, and trades \\
Research period & 2022--2023 \\
Out-of-sample period & Sealed year 2024 \\
\bottomrule
\end{tabular}
\end{table}

The data interface restricts access by date, instrument, and field. At each decision date, the strategy can access authorized historical observations, including permitted warmup history. The same input definitions and access rules apply during research and assessment.

\subsection{Research and Assessment Protocol}
\label{sec:protocol}

Models may evaluate candidate strategies on the research period and revise their code using aggregate return, signal-quality and coverage statistics, and execution errors. Each run may evaluate at most five candidates on the research period. Participants may define training and validation splits within the research period; the benchmark prescribes no separate training window.

At the end of research, one strategy's source code, execution environment, and declared features are frozen. The evaluator applies this strategy once to the sealed out-of-sample period. The strategy accesses the observations needed for execution through the authorized data interface. Neither these observations nor assessment feedback is returned to the model for further development.

\subsection{Financial Evaluation and Metrics}
\label{sec:financial}
\label{sec:scoring}

The evaluator converts stock rankings into an equal-weight long portfolio targeting 75 stocks. The backtest maintains a cash account with 100-share lots, order-dependent trading costs, and market impact estimated from a simulated order book. Idle cash earns zero return, and daily net asset value includes cash and outstanding positions. Account returns are compounded over each evaluation period.

We report out-of-sample annualized return and excess return over a full-market equal-weight reference aligned to the same trading days. The reference is gross of costs. Long-only Sharpe ($\mathrm{SR}_{\mathrm{L}}$) measures risk-adjusted performance, and maximum drawdown measures downside risk. Signal metrics include mean daily RankIC, its annualized information ratio ($A$), long--short Sharpe ($\mathrm{SR}_{\mathrm{LS}}$) between the highest- and lowest-scored deciles, and turnover, the average daily replacement fraction of the portfolio. Daily RankIC is the rank correlation between stock scores and next-period returns; $A$ is its mean divided by its standard deviation, multiplied by $\sqrt{N_{\mathrm{ann}}}$, where $N_{\mathrm{ann}}$ denotes the number of trading days per year used for annualization.

The published in-sample score combines strategy quality and feature diversity. Let $r$ be the strategy's annualized return minus the reference's annualized return, expressed as a decimal. Each metric is mapped to $(0,1)$ by a smooth, monotone transform,
\[
g_c(x)=\tfrac12\left(1+\frac{x}{c+|x|}\right),
\]
which equals $0.5$ at $x=0$, rewards gains with diminishing returns, and penalizes losses without truncating them to zero. The component scores are

\begin{equation}
\begin{aligned}
s_{\mathrm{R}}&=g_{0.3}(r), &
s_{\mathrm{L}}&=g_{1}(\mathrm{SR}_{\mathrm{L}}), &
s_{\mathrm{LS}}&=g_{3}(\mathrm{SR}_{\mathrm{LS}}),\\
s_{\mathrm{IC}}&=\tfrac12\left(1+\tanh\frac{\mathrm{RankIC}}{0.03}\right), &
s_{\mathrm{ST}}&=g_{2}(A), &
s_{\mathrm{TO}}&=\frac{1}{1+\mathrm{Turnover}}.
\end{aligned}
\label{eq:terms}
\end{equation}

Quality is a weighted sum that places most of its weight on realized excess return:

\begin{equation}
Q=0.6\,s_{\mathrm{R}}+0.2\,s_{\mathrm{L}}+0.05\left(s_{\mathrm{LS}}+s_{\mathrm{IC}}+s_{\mathrm{ST}}+s_{\mathrm{TO}}\right).
\label{eq:quality}
\end{equation}

Excess return and long-only Sharpe measure what the portfolio earned; the four signal terms carry small weights because they vary less across runs and do not by themselves determine portfolio value. A strategy that tracks the market with no excess return and zero Sharpe has $s_{\mathrm{R}}=s_{\mathrm{L}}=0.5$.

A strategy may declare up to 300 named features. Correlated features are grouped, and $K$ counts the qualifying clusters whose contribution to the strategy's scores is detected. Diversity is

\begin{equation}
D=Q\,\frac{\min(K,32)}{32}.
\label{eq:ed}
\end{equation}

The fixed scale of 32 makes diversity grow linearly with $K$ up to saturation, independently of the 300-feature declaration limit. Diversity scales with quality, so additional features add little to a weak strategy. The in-sample score is

\begin{equation}
S_{\mathrm{in}}=90\,Q+10\,D.
\label{eq:window}
\end{equation}

The out-of-sample score $S_{\mathrm{out}}$ uses the same metric dimensions with an unpublished combination rule fixed for each benchmark version. The overall score is

\begin{equation}
S=0.3\,S_{\mathrm{in}}+0.7\,S_{\mathrm{out}}.
\label{eq:total}
\end{equation}

We report individual metrics alongside the 0--100 scores.

\section{Experiments}
\label{sec:experiments}

\subsection{Experimental Setup}

Our experiments address five questions. To evaluate how far current models are from a strong research agent, and how reasoning effort changes this, we compare 38 iterative configurations of nine models (Section~\ref{sec:overall}). To evaluate why predictive skill does not become profit, we analyze next-day returns by signal decile (Section~\ref{sec:signal}). To evaluate whether any strategy adapts to the market regime, we trace Astra's account and exposure (Section~\ref{sec:timing}). To evaluate what iteration contributes, we pair iterative and one-shot runs and follow candidates through the research period (Section~\ref{sec:iteration}). To evaluate whether models share a trading style, we measure their holding exposures (Section~\ref{sec:style}).

We run \ModelClaudeOpus{}, \ModelClaudeSonnet{}, \ModelAstra{}, \ModelSol{}, \ModelTerra{}, \ModelGemini{}, \ModelDeepSeek{}, \ModelKimi{}, and \ModelGLM{} in the iterative and one-shot settings at every reasoning effort they support: six levels for Terra, five for Opus, Sonnet, Astra, and Sol, four for DeepSeek, three for Gemini and Kimi, and two for GLM. This gives 38 iterative and 38 one-shot configurations. The one-shot results are in Appendix~\ref{app:oneshot}. We report the total score $S$ with its research-period and sealed-period components, and for the sealed 2024 window the annualized return (ARR), excess ARR over the gross equal-weight market, mean RankIC, long-only Sharpe ratio, and maximum drawdown. The diagnostics in Sections~\ref{sec:signal}--\ref{sec:style} use, for each model, the iterative configuration with the highest total score (the shaded rows of Table~\ref{tab:main}).

\subsection{Performance Across Models and Reasoning Efforts}
\label{sec:overall}

\begin{table}[!t]
\centering\footnotesize
\caption{Iterative results for every model and supported reasoning effort, Models are ordered by their best total score, and efforts increase within each model. Shaded rows mark each model's highest-scoring effort, the configuration used by the diagnostics in Sections~\ref{sec:signal}--\ref{sec:style}. Scores are on a 0--100 scale. Financial and signal metrics refer to the sealed 2024 window; the gross equal-weight market ARR is 17.1\%. MDD is a negative return. Bold marks the best value in each column.}
\label{tab:main}
\setlength{\tabcolsep}{3.4pt}
\begin{tabular}{@{}llrrrrrrrr@{}}
\toprule
Model & Effort & $S$ & $S_{\mathrm{in}}$ & $S_{\mathrm{out}}$ & ARR (\%) & Excess (pp) & RankIC & Sharpe & MDD (\%) \\
\midrule
\modellogo{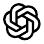}\ModelAstra & low & 47.22 & \textbf{58.10} & 42.56 & 2.3 & $-14.8$ & 0.056 & 0.22 & $-25.7$ \\
 & medium & 47.17 & 48.09 & 46.78 & 9.8 & $-7.2$ & 0.044 & 0.53 & $-17.9$ \\
 & high & 43.70 & 52.34 & 39.99 & $-0.9$ & $-17.9$ & 0.053 & 0.11 & $-26.3$ \\
\rowcolor{gray!14} & xhigh & \textbf{51.29} & 53.94 & 50.16 & 10.7 & $-6.3$ & \textbf{0.101} & 0.86 & $-6.8$ \\
 & max & 50.63 & 52.94 & 49.64 & 11.2 & $-5.9$ & 0.082 & 0.81 & $-7.0$ \\
\midrule
\modellogo{openai}\ModelSol & low & 38.51 & 48.85 & 34.09 & 1.4 & $-15.7$ & $-0.020$ & 0.19 & $-30.1$ \\
 & medium & 49.80 & 44.14 & \textbf{52.23} & \textbf{18.4} & \textbf{1.4} & 0.018 & 0.65 & $-35.9$ \\
 & high & 42.89 & 52.16 & 38.92 & $-2.5$ & $-19.5$ & 0.042 & 0.10 & $-37.4$ \\
\rowcolor{gray!14} & xhigh & 51.14 & 55.50 & 49.26 & 8.6 & $-8.4$ & 0.053 & \textbf{1.89} & $\mathbf{-2.8}$ \\
 & max & 45.31 & 46.49 & 44.81 & 8.1 & $-9.0$ & 0.018 & 0.41 & $-30.0$ \\
\midrule
\modellogo{openai}\ModelTerra & none & 33.09 & 26.67 & 35.85 & 2.4 & $-14.7$ & $-0.042$ & 0.22 & $-24.9$ \\
 & low & 41.19 & 50.58 & 37.17 & $-4.6$ & $-21.7$ & 0.022 & 0.02 & $-31.4$ \\
 & medium & 39.50 & 43.17 & 37.93 & $-4.3$ & $-21.3$ & 0.036 & 0.00 & $-29.0$ \\
 & high & 41.54 & 44.72 & 40.17 & 0.0 & $-17.0$ & 0.035 & 0.14 & $-29.9$ \\
 & xhigh & 44.77 & 50.60 & 42.28 & 9.9 & $-7.1$ & $-0.012$ & 0.44 & $-32.6$ \\
\rowcolor{gray!14} & max & 47.58 & 50.37 & 46.39 & 8.4 & $-8.7$ & 0.026 & 0.42 & $-28.8$ \\
\midrule
\modellogo{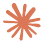}\ModelClaudeOpus & low & 30.15 & 37.76 & 26.89 & $-29.4$ & $-46.4$ & 0.035 & $-1.21$ & $-44.1$ \\
\rowcolor{gray!14} & medium & 44.28 & 56.78 & 38.92 & 4.4 & $-12.7$ & $-0.002$ & 0.29 & $-32.7$ \\
 & high & 39.89 & 52.87 & 34.32 & $-11.2$ & $-28.3$ & 0.039 & $-0.25$ & $-32.7$ \\
 & xhigh & 41.92 & 54.48 & 36.53 & $-7.0$ & $-24.0$ & 0.043 & $-0.08$ & $-33.2$ \\
 & max & 41.39 & 50.41 & 37.52 & $-4.0$ & $-21.1$ & 0.030 & $-0.03$ & $-25.1$ \\
\midrule
\modellogo{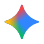}\ModelGemini & low & 41.46 & 53.44 & 36.32 & $-6.2$ & $-23.3$ & 0.037 & $-0.04$ & $-35.8$ \\
\rowcolor{gray!14} & medium & 43.04 & 53.58 & 38.52 & $-2.9$ & $-19.9$ & 0.045 & 0.06 & $-33.8$ \\
 & high & 35.39 & 48.90 & 29.61 & $-19.1$ & $-36.2$ & 0.022 & $-0.49$ & $-38.5$ \\
\midrule
\modellogo{claude}\ModelClaudeSonnet & low & 34.97 & 47.91 & 29.43 & $-9.5$ & $-26.6$ & $-0.009$ & $-0.13$ & $-35.5$ \\
 & medium & 39.42 & 43.39 & 37.73 & $-3.5$ & $-20.6$ & 0.041 & $-0.02$ & $-24.3$ \\
 & high & 40.50 & 54.11 & 34.67 & $-9.0$ & $-26.1$ & 0.044 & $-0.19$ & $-34.7$ \\
 & xhigh & 40.10 & 44.68 & 38.13 & $-0.5$ & $-17.6$ & 0.015 & 0.15 & $-35.0$ \\
\rowcolor{gray!14} & max & 40.93 & 57.33 & 33.90 & $-12.2$ & $-29.2$ & 0.045 & $-0.23$ & $-36.1$ \\
\midrule
\modellogo{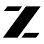}\ModelGLM & high & 38.67 & 49.31 & 34.11 & $-8.6$ & $-25.7$ & 0.018 & $-0.14$ & $-31.6$ \\
\rowcolor{gray!14} & max & 38.91 & 47.96 & 35.03 & $-7.8$ & $-24.9$ & 0.046 & $-0.18$ & $-27.8$ \\
\midrule
\rowcolor{gray!14}\modellogo{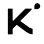}\ModelKimi & low & 38.31 & 45.87 & 35.07 & $-8.5$ & $-25.6$ & 0.027 & $-0.10$ & $-36.4$ \\
 & high & 31.43 & 36.85 & 29.10 & $-20.0$ & $-37.0$ & 0.036 & $-0.84$ & $-36.9$ \\
 & max & 31.86 & 39.87 & 28.43 & $-25.6$ & $-42.7$ & 0.041 & $-1.14$ & $-39.9$ \\
\midrule
\modellogo{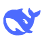}\ModelDeepSeek & none & 22.00 & 21.21 & 22.33 & $-24.1$ & $-41.2$ & $-0.022$ & $-1.03$ & $-29.5$ \\
\rowcolor{gray!14} & low & 36.45 & 48.46 & 31.30 & $-5.7$ & $-22.7$ & $-0.009$ & $-0.06$ & $-30.3$ \\
 & high & 35.97 & 46.02 & 31.67 & $-11.3$ & $-28.4$ & 0.015 & $-0.25$ & $-34.0$ \\
 & max & 35.90 & 42.49 & 33.08 & $-11.0$ & $-28.1$ & 0.037 & $-0.28$ & $-34.2$ \\
\bottomrule
\end{tabular}

\end{table}

\noindent\textbf{Profitable strategies are rare, and only one beats the market.} Thirteen of the 38 configurations earn positive ARR, and the only one above the 17.1\% market is Sol at medium effort (18.4\%, $+1.4$ pp). Most strategies keep under half of their capital in stocks (mean exposure 34--48\% in 35 of the 38 runs), so they pay trading costs while capturing only part of the market's gain.

\noindent\textbf{Higher reasoning effort does not reliably raise the score.} For six of the nine models the best effort is not the highest one supported (Table~\ref{tab:main}): Opus scores 44.28 at medium against 41.39 at max, Gemini 43.04 against 35.39, and Kimi 38.31 at low against 31.86 at max. Only Sonnet, Terra, and GLM peak at their highest effort. The range across a model's efforts reaches 14.49 points (Terra), close to the 18.77-point range between the nine highest-effort scores. Effort acts through discrete design choices rather than gradual improvement. Astra's xhigh and max runs add a market-state filter and hold cash on 70--75\% of days, while its low, medium, and high runs stay about 47--48\% invested, and its score jumps from 43.70 at high to 51.29 at xhigh. Sol follows the same pattern: its xhigh run is 5\% invested on average against 34--46\% at its other efforts.

\subsection{Why Models Fall Short: Ranking Skill Does Not Reach the Portfolio}
\label{sec:signal}

\begin{figure}[H]
\centering
\includegraphics[width=0.8\linewidth]{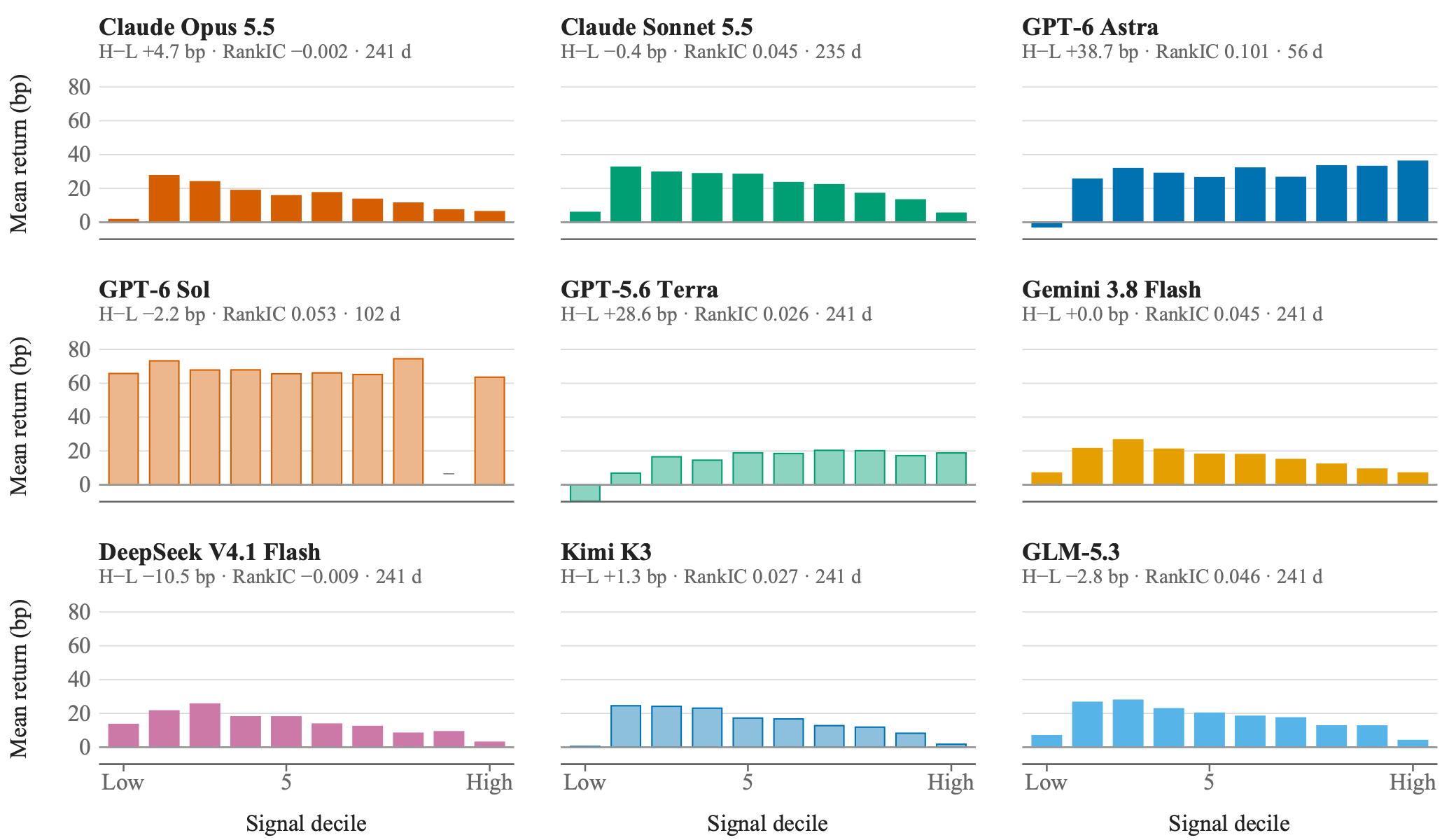}
\caption{Mean next-day return by daily score decile for the nine models' best iterative configurations in 2024, in basis points (bp). H--L is the highest-minus-lowest decile spread, and the last entry in each header is the number of days with a signal.}
\label{fig:deciles}
\end{figure}

\noindent\textbf{Models rank stocks with some skill, but the stocks they buy are not the ones that earn the most.} Opus, Sonnet, Gemini, DeepSeek, Kimi, and GLM show a hump (Figure~\ref{fig:deciles}): the second or third decile earns the most, 24--33 bp per day, and returns fall toward the top decile, so H--L lies between $-10.5$ and $+4.7$ bp. Terra and Astra rise across deciles ($+28.6$ and $+38.7$ bp), and Sol is flat. Average rank correlation does not predict profit: GLM has RankIC 0.046 and ARR $-7.8\%$, while Opus at medium effort has RankIC $-0.002$ and ARR 4.4\%, and across the 38 iterative runs the two are almost uncorrelated (Spearman 0.09). RankIC averages over the whole cross-section, but the portfolio holds only the 75 top-ranked stocks, and for most models that is where the decile profile declines.

\subsection{Case Study: Astra Learns Market Timing}
\label{sec:timing}

\begin{figure}[H]
\centering
\includegraphics[width=\linewidth]{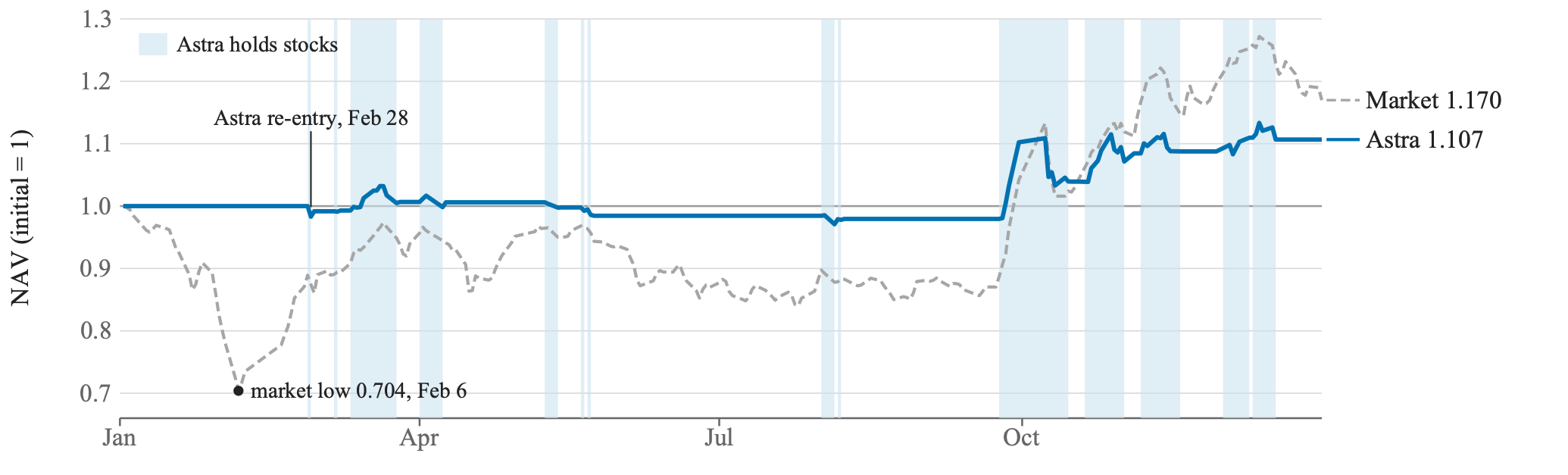}
\caption{Cash-account NAV of the iterative Astra run at xhigh effort in 2024, against the gross equal-weight market. Shaded bands mark days on which Astra holds stocks.}
\label{fig:astra}
\end{figure}

\noindent\textbf{Astra's lead comes from staying out of the market, not from choosing better stocks.} Its program scores stocks only when the median five-session and twenty-session log returns of the universe are both positive, and holds cash otherwise. The first three candidates have no such rule. It appears in the fourth and fifth submissions, the fifth being the committed strategy, after research feedback showed net losses in both research years ($-12.2\%$ in 2022 and $-13.6\%$ in 2023). The rule keeps Astra out of the early-2024 decline (Figure~\ref{fig:astra}): it first buys stocks on 28 February, its NAV is exactly 1.000 when the market bottoms at 0.704 on 6 February, and its maximum drawdown over January and February is $-1.7\%$ against $-29.6\%$ for the market. Over the year, Astra ends at 1.107 against 1.170 for the market, $-6.3$ pp of excess return, and 12.5\% of its gain arrives in the five sessions from 24 to 30 September. Cash on 75\% of trading days caps participation: Astra captures 16\% of the market's rebound from the February low, against 73\% for Terra, so the lower drawdown comes with a lower upside. The behavior depends on effort: Astra's low, medium, and high runs are invested on nearly every day and suffer drawdowns of $-25.7\%$, $-17.9\%$, and $-26.3\%$.

\subsection{Iteration Raises Scores More Reliably Than Returns}
\label{sec:iteration}

\begin{table}[!t]
\centering\footnotesize
\caption{Paired iterative and one-shot results, each at the supported effort with the highest total score within its setting (efforts listed as iterative / one-shot). $\Delta$ is iterative minus one-shot. $S_{\mathrm{in}}$ and $S_{\mathrm{out}}$ are the research-period and sealed-period scores, and ARR differences are in percentage points. The gross equal-weight market ARR is 17.1\%.}
\label{tab:paired}
\setlength{\tabcolsep}{4pt}
\begin{tabular}{@{}llrrrrrrrr@{}}
\toprule
Model & Efforts & \multicolumn{3}{c}{$S$} & \multicolumn{2}{c}{Window $\Delta$} & \multicolumn{3}{c}{ARR (\%)} \\
\cmidrule(lr){3-5}\cmidrule(lr){6-7}\cmidrule(lr){8-10}
 & & Iter. & One-shot & $\Delta$ & $\Delta S_{\mathrm{in}}$ & $\Delta S_{\mathrm{out}}$ & Iter. & One-shot & $\Delta$ \\
\midrule
\ModelAstra & xhigh / high & 51.29 & 35.22 & 16.07 & 15.67 & 16.24 & 10.7 & $-4.8$ & 15.5 \\
\ModelSol & xhigh / low & 51.14 & 32.15 & 18.99 & 17.21 & 19.74 & 8.6 & $-7.6$ & 16.2 \\
\ModelTerra & max / none & 47.58 & 42.14 & 5.44 & 3.81 & 6.14 & 8.4 & 9.5 & $-1.1$ \\
\ModelClaudeOpus & medium / medium & 44.28 & 42.61 & 1.67 & 9.59 & $-1.73$ & 4.4 & 5.6 & $-1.2$ \\
\ModelGemini & medium / medium & 43.04 & 35.91 & 7.13 & 17.96 & 2.48 & $-2.9$ & 1.6 & $-4.5$ \\
\ModelClaudeSonnet & max / high & 40.93 & 40.24 & 0.69 & 13.05 & $-4.60$ & $-12.2$ & 1.4 & $-13.6$ \\
\ModelGLM & max / high & 38.91 & 38.02 & 0.89 & $-4.58$ & 3.24 & $-7.8$ & $-4.2$ & $-3.6$ \\
\ModelKimi & low / high & 38.31 & 32.01 & 6.30 & 7.98 & 5.59 & $-8.5$ & $-20.5$ & 12.0 \\
\ModelDeepSeek & low / max & 36.45 & 29.92 & 6.53 & 12.51 & 3.96 & $-5.7$ & $-11.9$ & 6.2 \\
\midrule
Mean & & 43.55 & 36.47 & 7.08 & 10.36 & 5.67 & $-0.6$ & $-3.4$ & 2.9 \\
\bottomrule
\end{tabular}

\end{table}

\noindent\textbf{Iteration raises the score for every model but the return for only four.} Pairing each model's best iterative and one-shot configurations (Table~\ref{tab:paired}), the mean score rises from 36.47 to 43.55, with gains from 0.69 (Sonnet) to 18.99 (Sol). ARR is higher in the iterative run for Sol, Astra, Kimi, and DeepSeek (by 6.2 to 16.2 pp) and lower for the other five, so the mean gain is only 2.9 pp. Feedback improves the research window more than the sealed year: the mean gain in $S_{\mathrm{in}}$ is 10.36 points against 5.67 in $S_{\mathrm{out}}$, and the iterative best configurations lose 12.4 points between the two windows against 7.7 for one-shot. Part of what iteration learns is specific to 2022--2023.

\begin{figure}[H]
\centering
\includegraphics[width=0.8\linewidth]{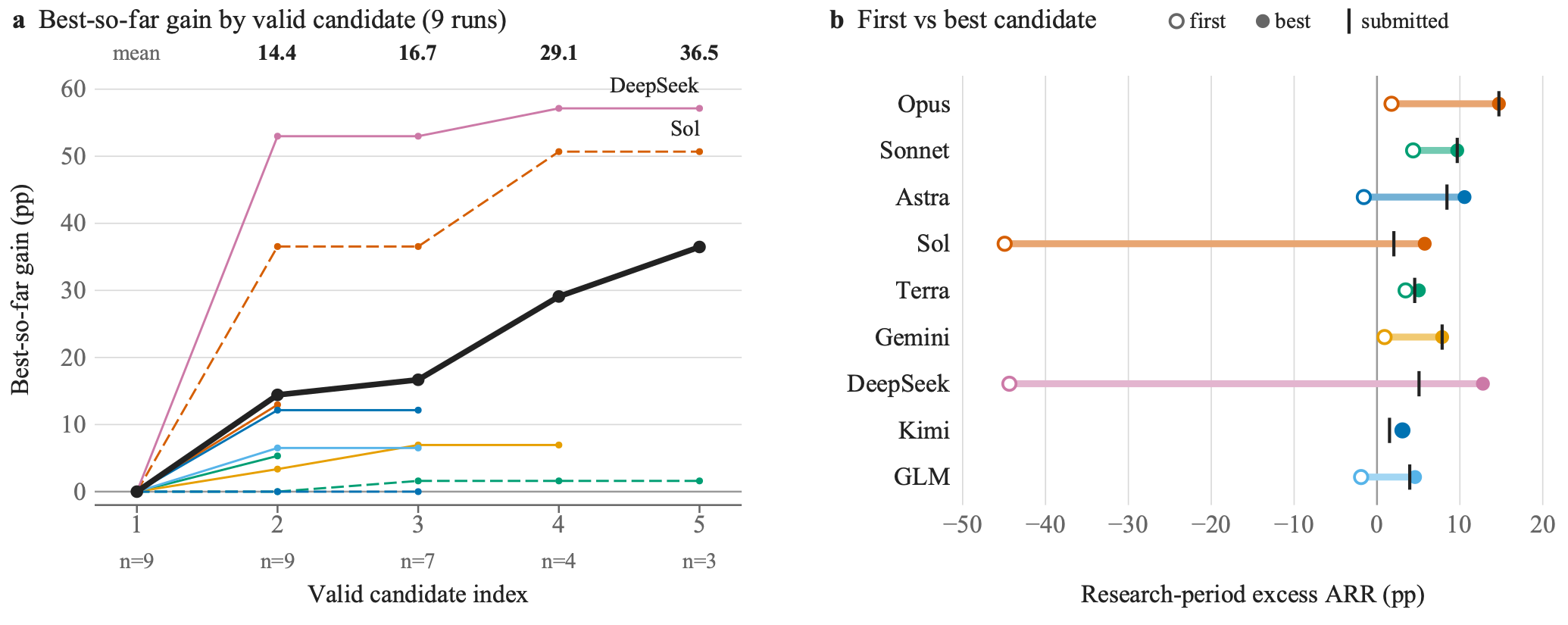}
\caption{Research-period progress of the nine iterative runs. (a)~Best-so-far excess ARR gain over the first valid candidate; means use the runs available at each index. (b)~First, best, and committed candidates on 2022--2023.}
\label{fig:progress}
\end{figure}

\noindent\textbf{Most research progress comes from repairing a weak first candidate.} Mean best-so-far gains at valid-candidate indices one to five are 0, 14.4, 16.7, 29.1, and 36.5 pp over 9, 9, 7, 4, and 3 runs (Figure~\ref{fig:progress}), but the median gain is 6.5 pp at the second candidate and 7.0 pp at the end of the run. Two runs drive the mean: DeepSeek and Sol improve by 57.1 and 50.7 pp from first candidates near $-44$ and $-45$ pp excess ARR, and without them the mean final gain is 6.5 pp. The second candidate is the research-period best in four of nine runs, and three runs commit their best candidate. Large gains come from fixing a clear error in a poor first candidate, and once candidates are near break-even, further evaluations add only a few points.

\subsection{Trading Styles}
\label{sec:style}

\begin{figure}[H]
\centering
\includegraphics[width=.9\linewidth]{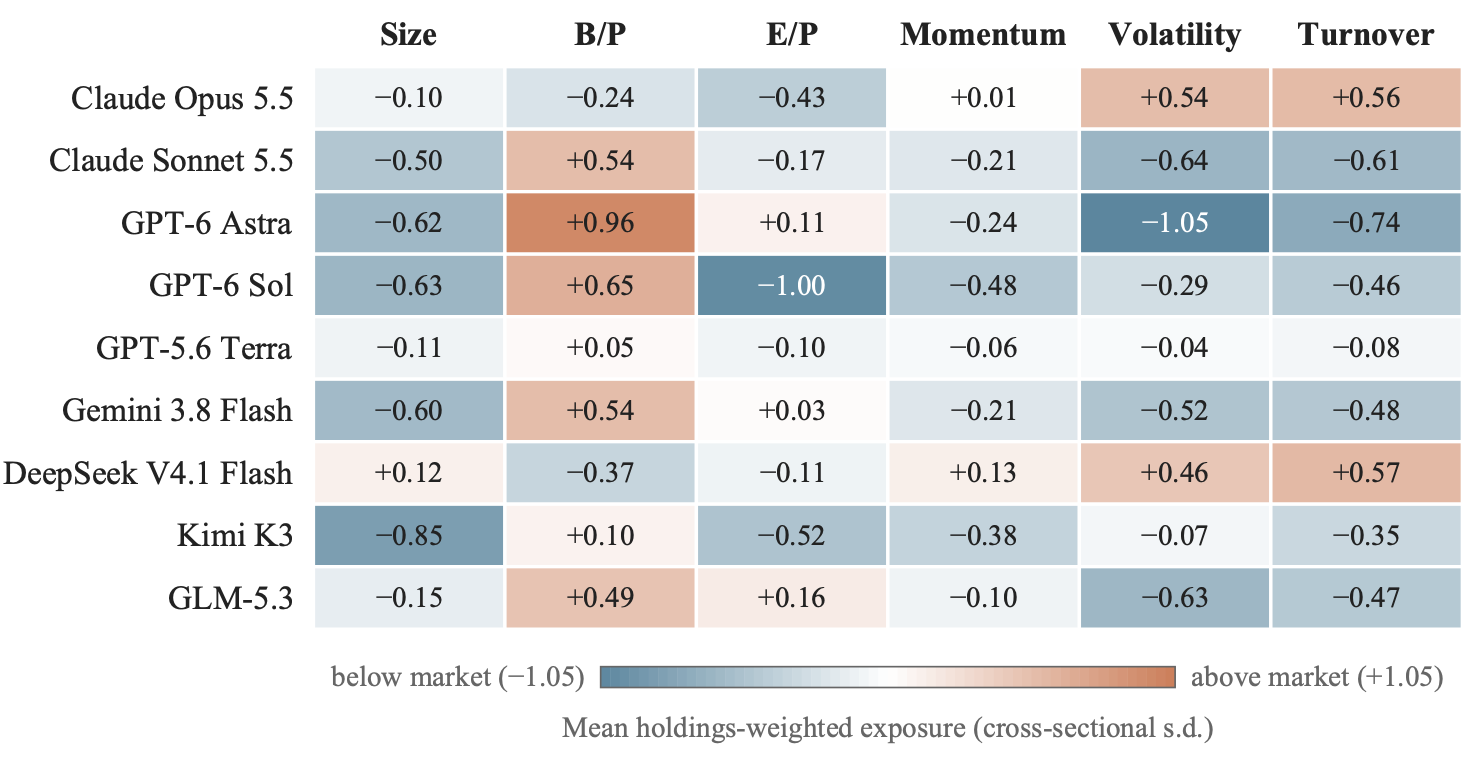}
\caption{Holding-value-weighted style exposure of the nine models' best iterative configurations in 2024, in cross-sectional standard deviations relative to the market. Astra's estimates use its 61 invested days.}
\label{fig:styles}
\end{figure}

\noindent\textbf{Most models converge on small-cap, low-volatility, value-tilted holdings, and style does not separate winners from losers.} Eight of nine models tilt toward smaller stocks, seven toward higher book-to-price, and most toward lower volatility and turnover (Figure~\ref{fig:styles}). Opus and DeepSeek are the exceptions, leaning toward higher volatility ($+0.54$ and $+0.46$) and turnover ($+0.56$ and $+0.57$), and Terra stays within 0.11 of the market on every dimension. Earnings-to-price is negative for six models, so the value tilt comes from book value rather than earnings. Similar tilts end in different places: Astra (book-to-price $+0.96$, volatility $-1.05$) finishes at 1.11 and Sonnet ($+0.54$, $-0.64$) at 0.88. Account outcomes are set by market exposure and by the head of each ranking, as the previous sections show, and the style tilt adds little to them.

\subsection{Kepler-1 Against the Models}
\label{sec:kepler}

\begin{figure}[H]
\centering
\includegraphics[width=\linewidth]{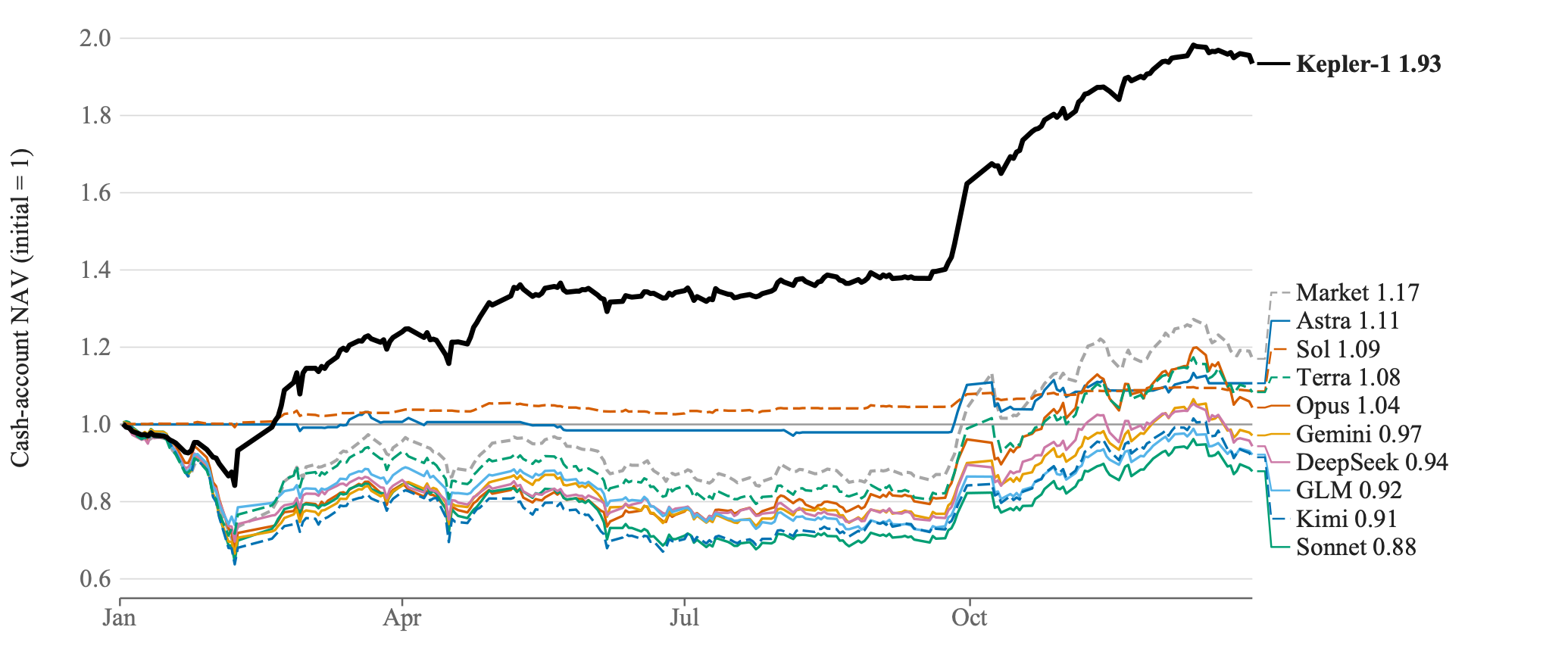}
\caption{Cash-account NAV in 2024 for the nine models' best iterative configurations and Kepler-1, against the gross equal-weight market.}
\label{fig:nav}
\end{figure}

\noindent\textbf{Kepler-1 turns its edge into account value, while the best model barely finishes above its starting capital.} Kepler-1 scores 77.56 and ends 2024 at 1.93 against 1.17 for the market (Figure~\ref{fig:nav}), and four models finish above their starting capital: Astra (1.11), Sol (1.09), Terra (1.08), and Opus (1.04). The other five end between 0.88 (Sonnet) and 0.97 (Gemini). The lead does not come from a better average rank signal: Kepler-1's RankIC of 0.085 is below Astra's 0.101. It comes from converting signal into return, with a Sharpe ratio of 2.98 against 1.89 for the best model.

\section{Related Work}
\label{sec:related}

\subsection{Formulaic Alpha Mining with Language Models}

Formulaic alpha mining searches for executable expressions that transform market observations into predictive cross-sectional signals. Published formula libraries demonstrate the interpretability and deployability of this representation \citep{Kakushadze2016Formulaic}. Automated approaches then use evolutionary search or learned generators to explore expression spaces and combine the resulting factors \citep{Zhang2020AutoAlpha,Shi2024AlphaForge}. These explicit search spaces support reproducible execution, but they are combinatorial, highly exposed to noisy backtest feedback, and often optimized around a narrow factor statistic rather than the behavior of the final portfolio.

LLMs extend this paradigm by connecting economic hypotheses, symbolic construction, and empirical diagnosis. LLMFactor uses sequential knowledge-guided prompting to extract interpretable factors from financial information \citep{Wang2024LLMFactor}; Automate Strategy Finding with LLM combines multimodal market inputs with a multi-agent portfolio workflow \citep{Kou2024AutomateStrategy}; Alpha-GPT introduces interactive LLM-based alpha generation \citep{Wang2023AlphaGPT}; AlphaAgent regularizes generation for originality, hypothesis--factor alignment, and structural simplicity \citep{Tang2025AlphaAgent}; and QuantaAlpha treats a complete mining trajectory as the unit of evolution and reuses effective sub-trajectories through mutation and crossover \citep{Han2026QuantaAlpha}. These systems establish the value of structured, multi-stage alpha discovery beyond one-shot formula generation. \BenchmarkName\ provides a common protocol for evaluating the resulting model runs with executable strategy outputs and a frozen final-assessment boundary.

\subsection{Quantitative Benchmarks and Evaluation Protocols}

Existing financial infrastructure serves several different evaluation goals. Qlib standardizes AI-oriented quantitative workflows, while FinRL provides reproducible reinforcement-learning environments and trading evaluations \citep{Yang2020Qlib,Liu2020FinRL}. Financial language benchmarks such as FinBen and Finance Agent Benchmark evaluate domain understanding, generation, forecasting, and real-world research tasks \citep{Xie2024FinBen,Bigeard2025FinanceAgent}; general agent benchmarks such as AgentBench and MLAgentBench evaluate multi-turn interaction, tool use, and iterative experimentation \citep{Liu2023AgentBench,Huang2023MLAgentBench}. AlphaBench focuses specifically on formulaic alpha generation, evaluation, and search tasks \citep{Luo2026AlphaBench}, while AlphaForgeBench targets reproducible end-to-end factor and strategy design with LLMs \citep{Zhang2026AlphaForgeBench}. Recent trading-agent evaluation also emphasizes leakage control, attribution, and memory effects in historical markets \citep{Zhu2026KTDFin}.

These benchmarks also differ in their unit of evaluation. AlphaQT-Bench~\citep{AlphaQTBench2026} treats executability, causal safety, functional correctness, and structural compliance as distinct code-level outcomes, with a small number of repair rounds. StockBench~\citep{StockBench2025} evaluates repeated sequential decisions on a small stock universe and reports passive-hold references, whereas QuantBench~\citep{QuantBench2025} studies conventional quantitative models, feature sets, and training seeds rather than LLM research systems. AlphaForgeBench expands the number of generated implementations through many short tasks, and KTD-Fin separates trading-agent behavior from information access and attribution. These designs motivate reporting execution status, financial outcomes, and information boundaries.

Evaluating a quantitative benchmark also requires controlling the process that produced the final output. Repeated access to noisy backtests creates opportunities for multiple testing and selection bias \citep{Bailey2017Backtest}. \BenchmarkName\ addresses this requirement through frozen data releases, date-scoped execution, bounded research budgets, bounded interaction feedback, explicit financial conventions, and sealed assessment of one committed strategy. The resulting protocol complements code, factor, and trading-decision benchmarks by connecting raw feature construction and bounded research to the sealed financial outcome of a committed strategy.

\section{Conclusion}

\BenchmarkName{} evaluates a complete quantitative research run, from constructing features on raw market observations to revising candidates and committing one strategy for sealed financial assessment. A shared stock-scoring interface, portfolio rule, and cash-account evaluator connect model-authored research to comparable simulated outcomes.

Across 38 iterative configurations, the best model scores 51.3 against 77.6 for Kepler-1, a proprietary strategy developed by Einsia AI, and only one configuration, Sol at medium effort, beats the equal-weight market. Pairing each model's best iterative and one-shot configurations, iteration raises the overall score for all nine models but sealed-period ARR for only four. Decile returns show why: most models rank stocks with some skill, yet the top-ranked names that the portfolio buys earn less than middle-ranked ones, so an average predictive correlation is insufficient to judge an investable strategy. The best-performing Astra configuration profits mainly by staying in cash during the early-2024 decline, which lowers drawdown without beating the market. Candidate trajectories show that research progress comes early and plateaus, and the effort sweep shows that a higher reasoning setting does not consistently improve the result. Together, these findings support evaluating quantitative research through both its development process and the financial outcome of its committed strategy.















\clearpage
\newpage
\bibliographystyle{assets/plainnat}
\bibliography{ref}


\clearpage
\newpage
\appendix

\section{Suite Specification}
\label{app:registry}

Table~\ref{tab:suite_registry} lists the benchmark specification. A strategy may declare up to 300 named features.

\begin{table}[htbp]
\centering\small
\caption{Public specification of the selected A-share suite.}
\label{tab:suite_registry}
\begin{tabular}{@{}p{0.24\linewidth}p{0.70\linewidth}@{}}
\toprule
Field & Specification \\
\midrule
Market and universe & Selected A-shares; static universe of 3,042 stocks; missing and delisted stocks retained \\
Input profile & Raw daily and Level-2 market data through a date-scoped data interface; Python strategies \\
Windows & In-sample research 2022--2023 with authorized warmup history; sealed out-of-sample 2024; participants may define training and validation splits within the research period \\
Budget & At most 5 research evaluations \\
Feedback (iterative track) & Aggregate return, signal, and coverage statistics on the research window, plus execution errors \\
Portfolio & Top-75 equal-weight long portfolio, boundary ties retained; Top-200/400 diagnostics \\
Accounting & Cash account with 100-share lots; idle cash earns zero; liquidation at window end \\
Costs & Order-dependent trading costs and market impact estimated from a simulated order book \\
Features & Up to 300 declared features; usage tested by score-rank regression on correlation clusters; diversity normalized by a fixed scale of 32 \\
Scoring & $S_{\mathrm{in}}=90Q+10D$; total $0.3\,S_{\mathrm{in}}+0.7\,S_{\mathrm{out}}$ \\
Runs & One run per submission; results published automatically and not withdrawable; daily submission limit per account \\
\bottomrule
\end{tabular}
\end{table}

\paragraph{Boards.} One-shot and Iterative use the standard model harness. System entries use participant-controlled agents; Custom entries support self-hosted models or unsupported endpoints with self-reported configurations.

\paragraph{Result records.} For every run, the platform records the model, track, harness, all model calls, every research evaluation, the committed strategy, and both window assessments. Failed and cancelled runs remain in the record. Each official leaderboard entry links to its run record.

\section{One-Shot Results and Matched-Effort Comparison}
\label{app:oneshot}

\subsection{One-Shot Results Across Reasoning Efforts}

\begin{table}[!t]
\centering\footnotesize
\caption{One-shot results for every model and supported reasoning effort, Layout and shading follow Table~\ref{tab:main}; shaded rows mark each model's highest-scoring one-shot effort. The one-shot setting receives no performance feedback.}
\label{tab:oneshot}
\setlength{\tabcolsep}{3.4pt}
\begin{tabular}{@{}llrrrrrrrr@{}}
\toprule
Model & Effort & $S$ & $S_{\mathrm{in}}$ & $S_{\mathrm{out}}$ & ARR (\%) & Excess (pp) & RankIC & Sharpe & MDD (\%) \\
\midrule
\modellogo{claude}\ModelClaudeOpus & low & 35.04 & 43.21 & 31.53 & $-4.5$ & $-21.6$ & $-0.011$ & 0.00 & $-33.0$ \\
\rowcolor{gray!14} & medium & \textbf{42.61} & 47.19 & \textbf{40.65} & 5.6 & $-11.5$ & $-0.001$ & 0.34 & $\mathbf{-21.0}$ \\
 & high & 36.07 & 39.68 & 34.52 & $-7.5$ & $-24.5$ & 0.020 & $-0.14$ & $-29.7$ \\
 & xhigh & 39.13 & 49.05 & 34.88 & $-9.9$ & $-26.9$ & 0.040 & $-0.20$ & $-35.0$ \\
 & max & 39.70 & 42.07 & 38.69 & $-3.7$ & $-20.8$ & \textbf{0.046} & 0.03 & $-35.9$ \\
\midrule
\rowcolor{gray!14}\modellogo{openai}\ModelTerra & none & 42.14 & 46.56 & 40.25 & \textbf{9.5} & $\mathbf{-7.6}$ & $-0.010$ & \textbf{0.44} & $-27.8$ \\
 & low & 27.82 & 26.59 & 28.35 & $-10.3$ & $-27.4$ & $-0.007$ & $-0.24$ & $-25.7$ \\
 & medium & 21.01 & 20.44 & 21.25 & $-32.4$ & $-49.4$ & $-0.010$ & $-1.51$ & $-34.1$ \\
 & high & 40.61 & 42.88 & 39.64 & 6.2 & $-10.9$ & $-0.001$ & 0.35 & $-27.5$ \\
 & xhigh & 34.21 & 42.94 & 30.46 & $-9.8$ & $-26.8$ & 0.000 & $-0.17$ & $-33.6$ \\
 & max & 20.98 & 20.98 & 20.97 & $-32.3$ & $-49.3$ & $-0.009$ & $-1.60$ & $-32.6$ \\
\midrule
\modellogo{claude}\ModelClaudeSonnet & low & 37.75 & 42.41 & 35.76 & 1.1 & $-16.0$ & $-0.007$ & 0.19 & $-30.3$ \\
 & medium & 33.82 & 41.81 & 30.39 & $-11.5$ & $-28.6$ & 0.002 & $-0.21$ & $-35.5$ \\
\rowcolor{gray!14} & high & 40.24 & 44.28 & 38.50 & 1.4 & $-15.6$ & 0.010 & 0.20 & $-32.1$ \\
 & xhigh & 37.05 & 47.90 & 32.40 & $-14.3$ & $-31.3$ & 0.029 & $-0.33$ & $-36.0$ \\
 & max & 33.89 & 41.19 & 30.77 & $-14.8$ & $-31.9$ & 0.018 & $-0.33$ & $-36.9$ \\
\midrule
\rowcolor{gray!14}\modellogo{glm}\ModelGLM & high & 38.02 & \textbf{52.54} & 31.79 & $-4.2$ & $-21.3$ & $-0.023$ & 0.05 & $-37.3$ \\
 & max & 31.46 & 34.70 & 30.07 & $-15.4$ & $-32.4$ & 0.022 & $-0.51$ & $-32.1$ \\
\midrule
\modellogo{gemini}\ModelGemini & low & 27.75 & 21.26 & 30.53 & $-12.1$ & $-29.1$ & 0.005 & $-0.41$ & $-23.5$ \\
\rowcolor{gray!14} & medium & 35.91 & 35.62 & 36.04 & 1.6 & $-15.4$ & $-0.006$ & 0.19 & $-23.8$ \\
 & high & 30.26 & 25.93 & 32.11 & $-5.4$ & $-22.5$ & $-0.003$ & $-0.06$ & $-24.6$ \\
\midrule
\modellogo{openai}\ModelAstra & low & 30.08 & 35.05 & 27.95 & $-11.2$ & $-28.3$ & $-0.010$ & $-0.28$ & $-30.3$ \\
 & medium & 31.02 & 36.73 & 28.57 & $-9.7$ & $-26.7$ & $-0.012$ & $-0.20$ & $-30.3$ \\
\rowcolor{gray!14} & high & 35.22 & 38.27 & 33.92 & $-4.8$ & $-21.9$ & 0.007 & $-0.03$ & $-30.9$ \\
 & xhigh & 30.30 & 38.46 & 26.80 & $-16.7$ & $-33.8$ & $-0.002$ & $-0.48$ & $-34.7$ \\
 & max & 31.47 & 40.13 & 27.75 & $-18.3$ & $-35.3$ & 0.007 & $-0.54$ & $-34.8$ \\
\midrule
\rowcolor{gray!14}\modellogo{openai}\ModelSol & low & 32.15 & 38.29 & 29.52 & $-7.6$ & $-24.7$ & $-0.014$ & $-0.08$ & $-32.2$ \\
 & medium & 22.60 & 22.62 & 22.59 & $-31.4$ & $-48.4$ & $-0.006$ & $-1.31$ & $-34.1$ \\
 & high & 23.15 & 21.07 & 24.04 & $-17.4$ & $-34.4$ & $-0.020$ & $-0.51$ & $-29.6$ \\
 & xhigh & 23.29 & 23.84 & 23.05 & $-19.0$ & $-36.1$ & $-0.022$ & $-0.60$ & $-29.6$ \\
 & max & 22.31 & 24.08 & 21.55 & $-23.7$ & $-40.7$ & $-0.018$ & $-0.98$ & $-32.5$ \\
\midrule
\modellogo{kimi}\ModelKimi & low & 31.58 & 38.58 & 28.58 & $-13.4$ & $-30.5$ & 0.000 & $-0.34$ & $-33.8$ \\
\rowcolor{gray!14} & high & 32.01 & 37.89 & 29.48 & $-20.5$ & $-37.6$ & 0.038 & $-0.84$ & $-38.6$ \\
 & max & 31.65 & 40.87 & 27.70 & $-20.3$ & $-37.3$ & 0.016 & $-0.86$ & $-35.1$ \\
\midrule
\modellogo{deepseek}\ModelDeepSeek & none & 20.48 & 22.01 & 19.82 & $-34.1$ & $-51.1$ & $-0.015$ & $-1.79$ & $-37.5$ \\
 & low & 29.36 & 34.21 & 27.28 & $-17.4$ & $-34.4$ & 0.001 & $-0.49$ & $-33.3$ \\
 & high & 29.63 & 37.51 & 26.25 & $-25.4$ & $-42.4$ & 0.010 & $-0.92$ & $-39.5$ \\
\rowcolor{gray!14} & max & 29.92 & 35.95 & 27.34 & $-11.9$ & $-29.0$ & $-0.008$ & $-0.42$ & $-26.7$ \\
\bottomrule
\end{tabular}

\end{table}

Table~\ref{tab:oneshot} reports the 38 one-shot runs in the layout of Table~\ref{tab:main}. The median configuration scores 31.61, and 4 of the 38 exceed 40 points, against 21 in the iterative setting. Six configurations have positive sealed-period ARR, and none exceeds the 17.1\% market.

\noindent\textbf{Without feedback, reasoning effort is just as inconsistent as with it.} For eight of the nine models the best effort is not the highest one supported (Table~\ref{tab:oneshot}). Terra scores 42.14 with no reasoning and 20.98 at max, a 21.16-point range that exceeds every iterative range, and Sol is best at low (32.15) and lowest at max (22.31). Only DeepSeek rises with effort, from 20.48 with no reasoning to 29.92 at max. With no feedback loop to correct a poor draft, each cell reflects a single draft, so the scores swing more with effort.

\subsection{Matched-Effort Comparison}

\noindent\textbf{Iteration helps at matched efforts as well.} At each model's highest supported effort (Gemini: high; others: max), the iterative run outscores the one-shot run for all nine models, with mean scores of 40.88 against 30.18, and has the higher ARR in six of the nine pairs. Across all 38 matched model--effort cells, the iterative run has the higher score in 34, the higher ARR in 26, and the higher RankIC in 28. Matching efforts gives a larger gap than pairing best configurations (Table~\ref{tab:paired}) because one-shot scores fall with effort more often, so a one-shot run at max effort is rarely its model's best.

\section{Additional Strategy Diagnostics}
\label{app:analysis}

\subsection{Signal Similarity and Portfolio Overlap}

\begin{figure}[H]
\centering
\includegraphics[width=\linewidth]{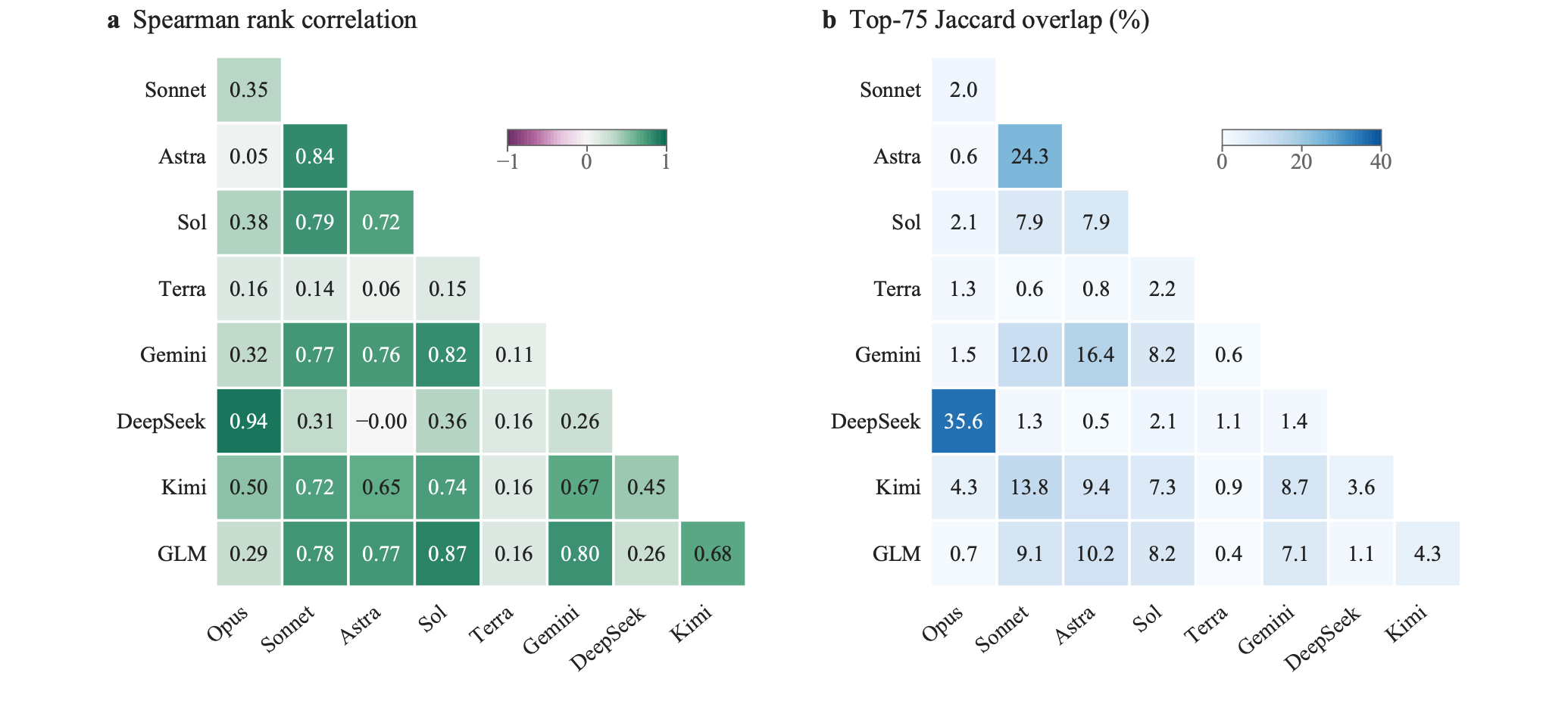}
\caption{Pairwise mean daily Spearman signal correlations (a) and Top-75 Jaccard overlaps (b) among the nine models' best iterative configurations in 2024 (36 pairs). Portfolio overlap measures the selected score sets, rather than identical realized trades under execution constraints.}
\label{fig:similarity}
\end{figure}

The mean pairwise rank correlation is 0.47, while mean Top-75 Jaccard overlap is only 6.1\% and its maximum is 35.6\% (Figure~\ref{fig:similarity}). Broadly similar rankings therefore need not select the same extreme tail. Opus and DeepSeek are the most similar pair, with correlation 0.94 and overlap 35.6\%, followed by Sol--GLM (0.87) and Sonnet--Astra (0.84). Terra is the most distinct strategy, with a mean correlation of 0.14 against the other eight.

Opus and DeepSeek are close to each other and correlate at about 0.5 or less with every other model. This agrees with the shared tilt toward higher volatility and turnover in Figure~\ref{fig:styles}. The remaining models largely form one cluster of related signals.

\end{document}